\documentclass{article}
\usepackage[T1]{fontenc}
\usepackage{xcolor}
\usepackage[final]{corl_2026} 

\usepackage{amsmath}
\usepackage{multirow}
\usepackage{float}
\usepackage{booktabs}
\usepackage{colortbl}
\usepackage{arydshln}
\usepackage{subcaption}
\usepackage{amssymb}
\usepackage{todonotes}
\usepackage[export]{adjustbox}

\newlength{\platformht}
\newlength{\envimght}

\newcommand{\ourmethod}{\texttt{OptCar}}

\title{Adapting Generalist Vehicle Models for \\High-Speed MPC Across Terrains}

\author{
  \textbf{Rwik Rana\textsuperscript{1},
  Jesse Quattrociocchi\textsuperscript{1,2},
  Christian Ellis\textsuperscript{1,2}},\\
  \textbf{Nathan Tsoi\textsuperscript{1},
  Garrett Warnell\textsuperscript{1,2},
  Joydeep Biswas\textsuperscript{1}}\\[2pt]
  \textsuperscript{1}The University of Texas at Austin
  \qquad
  \textsuperscript{2}DEVCOM Army Research Laboratory\\
  \texttt{rwik2000@utexas.edu}
}

\begin{document}
\maketitle

\begingroup
\renewcommand{\thefootnote}{}
\footnotetext{Project page: \url{https://amrl.cs.utexas.edu/optcar/}}
\endgroup

\begin{abstract}
High-speed off-road autonomy requires precise closed-loop control for a target vehicle while remaining robust across changing terrains. Recent forward kinodynamic (FKD) prediction foundation models suggest a promising path, starting from a generalist model and specializing it to the target platform. However, effective specialization remains challenging, as it often requires substantial real-world data, and models adapted to one setting can still overfit to specific terrains or driving regimes. We present \ourmethod{} (Optimized Car), a recipe for bridging the gap from generalist to specialist FKD models that preserves cross-terrain generalization while optimizing performance for a specific vehicle. \ourmethod{} introduces a transformer FKD architecture that uses FiLM to condition multi-step predictions on a single dynamics context token summarizing recent state-action history. It then specializes the generalist model using limited real-world data and targeted synthetic rollouts from environment-specific system identification. In closed-loop model predictive control (MPC) experiments across three terrains and an out-of-distribution cart-pulling task, the largest gains appear at 6~m/s, the highest speed evaluated and the regime in which slip dominates tracking error. On vegetation + dirt, the most slip-diverse terrain, \ourmethod{} reduces 6~m/s trajectory tracking error by roughly 55\% relative to AnyCar fine-tuned on real data alone, and remains the most accurate even when an unseen cart payload changes the dynamics. With 5 minutes of real data per terrain, \ourmethod{} is competitive on road with a specialist trained on 30 minutes of road data and outperforms it when the terrain changes.
\end{abstract}

\vspace{-8pt}

\section{Introduction}
\vspace{-8pt}

\begin{figure}[t]
    \centering
    \includegraphics[width=\columnwidth]{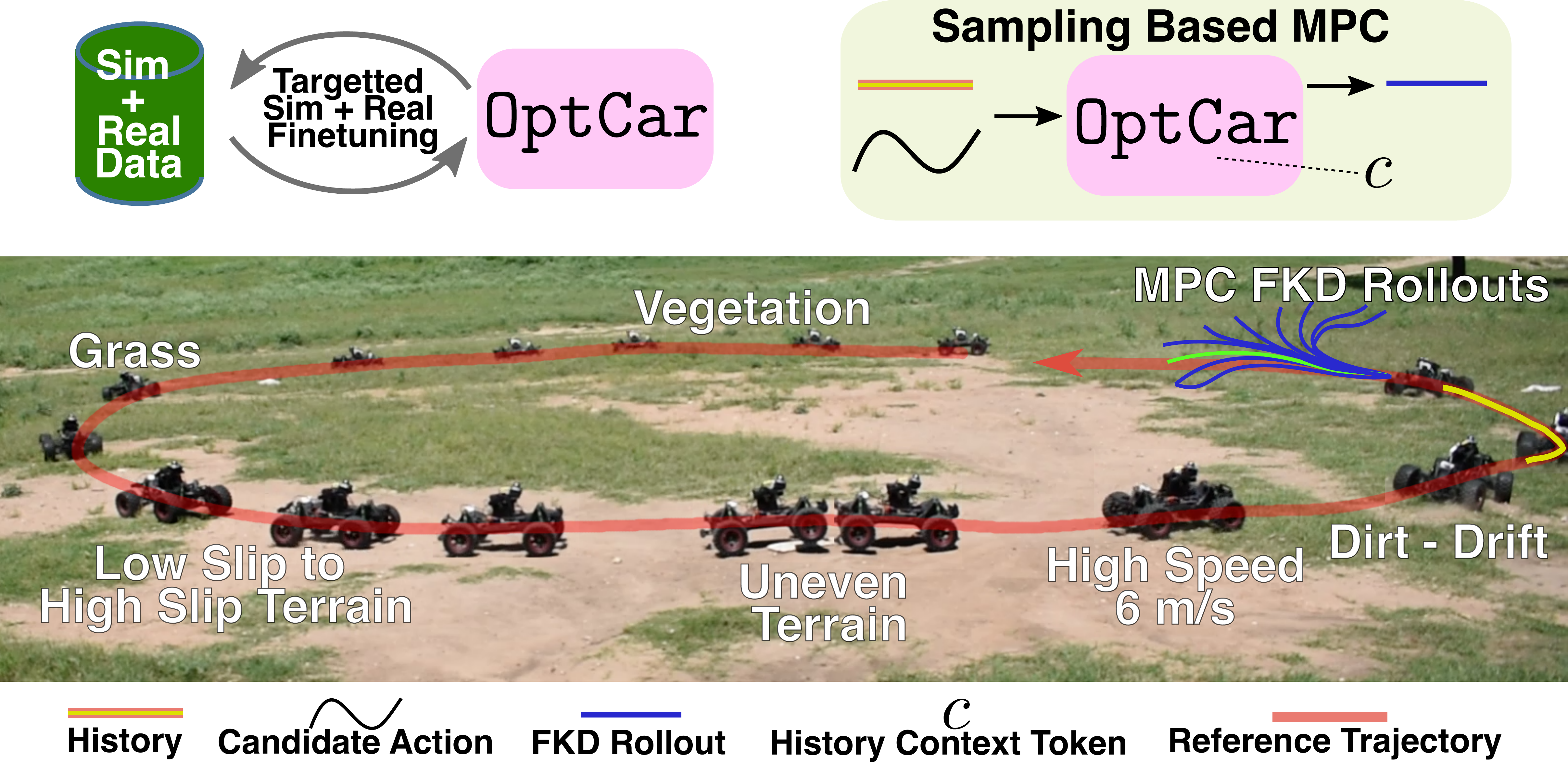}
    \caption{\textbf{\ourmethod{} driving at 6~m/s across mixed terrain.} The vehicle tracks a reference trajectory through a mixture of grass, vegetation, and dirt using sampling-based MPC. \ourmethod{} specializes a generalist FKD model through two contributions: a history context vector $\mathbf{c}$ that conditions the rollout decoder through FiLM to adapt to the current vehicle-terrain interaction, and a targeted real-and-synthetic fine-tuning recipe. \vspace{-12pt}}
    \label{fig:title}
\end{figure}

High-speed driving over unstructured terrain underlies applications from search-and-rescue and agricultural robotics to autonomous off-road racing. At high speed, the tires operate near their friction limit, and the resulting slip varies with terrain and payload, so a controller must be both precise on the target vehicle and robust as the terrain changes. Analytical models such as the dynamic bicycle model are fast, but simplified vehicle and tire dynamics can limit prediction accuracy~\cite{rajamani2011vehicle,raji2023tricycle}, motivating learned models and corrections~\cite{kabzan2019learning}. To run in real time on embedded hardware at high speed the model must be small, so accuracy must come under a tight fidelity-latency budget rather than from a larger network.

Forward kinodynamic (FKD) models predict future states from candidate action sequences and serve directly as the rollout model inside sampling-based model predictive control (MPC), such as MPPI~\cite{MPPI_2015}, coupling planning and control. Generalist FKD models such as AnyCar~\cite{anycar} generalize across vehicles and environments but lack the per-vehicle accuracy needed for high-speed control, whereas inverse kinodynamic (IKD) methods~\cite{imuikd, viikd} adapt online to the current terrain through an explicit history component, but rely on a separate planner. At the other extreme, specialist models fit a single vehicle and terrain from real data~\cite{williams2017ITMPC, drift_dojo}, achieving aggressive performance in that setting, while transfer to new terrains may require further adaptation. Specializing a generalist FKD model to a known vehicle while remaining robust across terrains is the central problem we address.

We present \ourmethod{} (Figure~\ref{fig:title}), which addresses these limitations and is validated by placing the model inside an MPPI controller~\cite{MPPI_2015} for closed-loop trajectory tracking. Its contributions are:
\begin{enumerate}
    \setlength{\itemsep}{1pt}
    \setlength{\parskip}{1pt}
    \item \textbf{A targeted real-and-synthetic fine-tuning recipe} that specializes the model to a target vehicle from only minutes of real data per terrain, augmenting it with synthetic rollouts from a per-terrain system-identified bicycle model to cover the high-slip regimes. This lowers 6~m/s tracking error by 17\% to 46\% over fine-tuning on real data alone.
    \item \textbf{FiLM-based history conditioning for FKD prediction} that compresses recent state-action history into a single dynamics context token and uses FiLM to condition every rollout decoder block, rather than cross-attending to individual history tokens. With \ourmethod{} and AnyCar, the history cross-attention baseline, both fine-tuned on the same real-and-synthetic dataset, \ourmethod{} reduces trajectory tracking error at 6~m/s by 19\%--30\%.
    \item \textbf{Closed-loop validation against generalist and specialist baselines.} Against AnyCar fine-tuned on real data alone, \ourmethod{} fine-tuned on real and synthetic data cuts 6~m/s tracking error by 42\% to 57\%. Against a specialist trained on 30 minutes of road data, it remains within 14\% on road yet tracks 46\% more accurately on unseen terrain. Under an unseen cart payload, it attains the lowest tracking error at every speed regime.
\end{enumerate}
\vspace{-10 pt}
\section{Related Work}
\vspace{-8 pt}
We review learned dynamics models for vehicle control, online dynamics adaptation, and data-efficient specialization via sim-to-real.
\vspace{-6 pt}

\subsection{Learned Dynamics Models for Vehicle Control}
\vspace{-10 pt}

We frame high-speed vehicle control as model-based planning under dynamics that vary with vehicle, terrain, and payload. Analytical kinematic and dynamic bicycle models~\cite{rajamani2011vehicle,raji2023tricycle} are fast and interpretable, but simplified vehicle and tire dynamics can limit prediction accuracy, while end-to-end policies~\cite{bojarski2016end} forgo the predictive rollouts needed to evaluate alternative futures. Learned forward kinodynamic (FKD) models occupy the middle ground we adopt, predicting future trajectories from candidate actions and serving as rollout models inside sampling-based MPC~\cite{MPPI_2015,williams2017ITMPC,Gandhi_2021}. Within FKD-MPC, generalist models such as AnyCar~\cite{anycar} train across many vehicles and environments for broad transfer, whereas specialist methods fit a single vehicle, terrain, or regime from real data: drifting and tire models~\cite{drift_dojo}, off-road and terrain-aware rollouts~\cite{gibson2023multistep,Lee_2023}, and information-theoretic or residual-dynamics racing controllers~\cite{williams2017ITMPC,kabzan2019learning,Xue_2024,kalaria2023adaptive}, while transfer to new terrains may require further adaptation. Our setting lies between the two, with a fixed vehicle that makes specialization worthwhile and a varying terrain that makes single-terrain specialization insufficient. 

\vspace{-8 pt}
\subsection{Online Dynamics Adaptation}
\vspace{-4pt}
Online adaptation methods use recent experience to account for changes in system dynamics, using state-action histories to condition policies~\cite{yu2017preparingunknownlearninguniversal,kumar2021rma}, represent changing disturbances~\cite{neuralfly,datt}, or perform terrain-adaptive ground-vehicle tracking~\cite{imuikd,viikd}. For learned dynamics models, prior work adapts through online parameter updates~\cite{levy2025metalearningonlinedynamicsmodel,tsuchiya2024onlineadaptationlearnedvehicle} or history-derived latent context for next-state prediction, as in CaDM~\cite{cadm2020}. Wang et al.~\cite{wang2023payattentiondrivesafe} use a transformer-derived context to condition a probabilistic LSTM dynamics model rolled forward within MPPI, while HyperMPC~\cite{hypermpc2025} predicts time-varying dynamics-model parameters from history and expected future controls. \ourmethod{} adapts in a single forward pass: a single history-derived dynamics context token modulates every block of a multi-step transformer FKD decoder through FiLM, directly shaping the candidate-action rollouts MPC evaluates.
\vspace{-8 pt}
\subsection{Data-Efficient Specialization and Sim-to-Real}
\vspace{-4pt}
\ourmethod{}'s second contribution, a targeted real-and-synthetic fine-tuning recipe, relates to the sim-to-real literature that uses limited real data to calibrate simulators or domain-randomization distributions~\cite{zhu2018fastmodelidentificationphysics, ramos2019bayessim, chebotar2018simopt, Muratore_2021, Dikici_2025}. These methods typically calibrate a simulator and then train or control within it, or tune a global randomization distribution. \ourmethod{} instead uses a per-terrain calibrated dynamic bicycle model purely as a targeted data generator that amplifies scarce real data into the high-slip regimes it under-samples; the learned FKD model, not the calibrated simulator, remains the rollout model deployed in MPC. This keeps the synthetic data anchored to the target vehicle rather than spread across broad randomization.
\vspace{-8pt}

\section{Problem Formulation}
\label{sec:problem}
\vspace{-8pt}
We consider forward kinodynamic (FKD) prediction for a target vehicle: given a history of recent states and applied actions together with a candidate future action sequence, predict the resulting future state trajectory. Let $\mathbf{s}_t \in \mathcal{S}$ denote the vehicle state (pose and velocity) at time $t$ and $\mathbf{a}_t \in \mathcal{A}$ the applied action (body-frame velocity and steering command). Given a history spanning indices $t-M$ through $t$ ($M+1$ entries) and prediction horizon $N$, we seek a deterministic prediction model $f_\theta$ that maps the recent history and a candidate future action sequence to the predicted future trajectory,
\begin{equation}
    \hat{\mathbf{s}}_{t+1:t+N} = f_\theta\!\left(\mathbf{s}_{t-M:t},\; \mathbf{a}_{t-M:t+N-1}\right),
    \label{eq:fkd}
\end{equation}
with parameters $\theta$ chosen to minimize prediction error over the predicted future states $\hat{\mathbf{s}}_{t+1:t+N}$ on the operational distribution of the target vehicle across its terrains. Because it runs in closed-loop control, the model must be small enough to roll out many candidate actions in real time.
\vspace{-10pt}

\section{Methodology}
\label{sec:method}
\vspace{-8pt}

\begin{figure}[t]
    \centering
    \includegraphics[width=1.0\columnwidth]{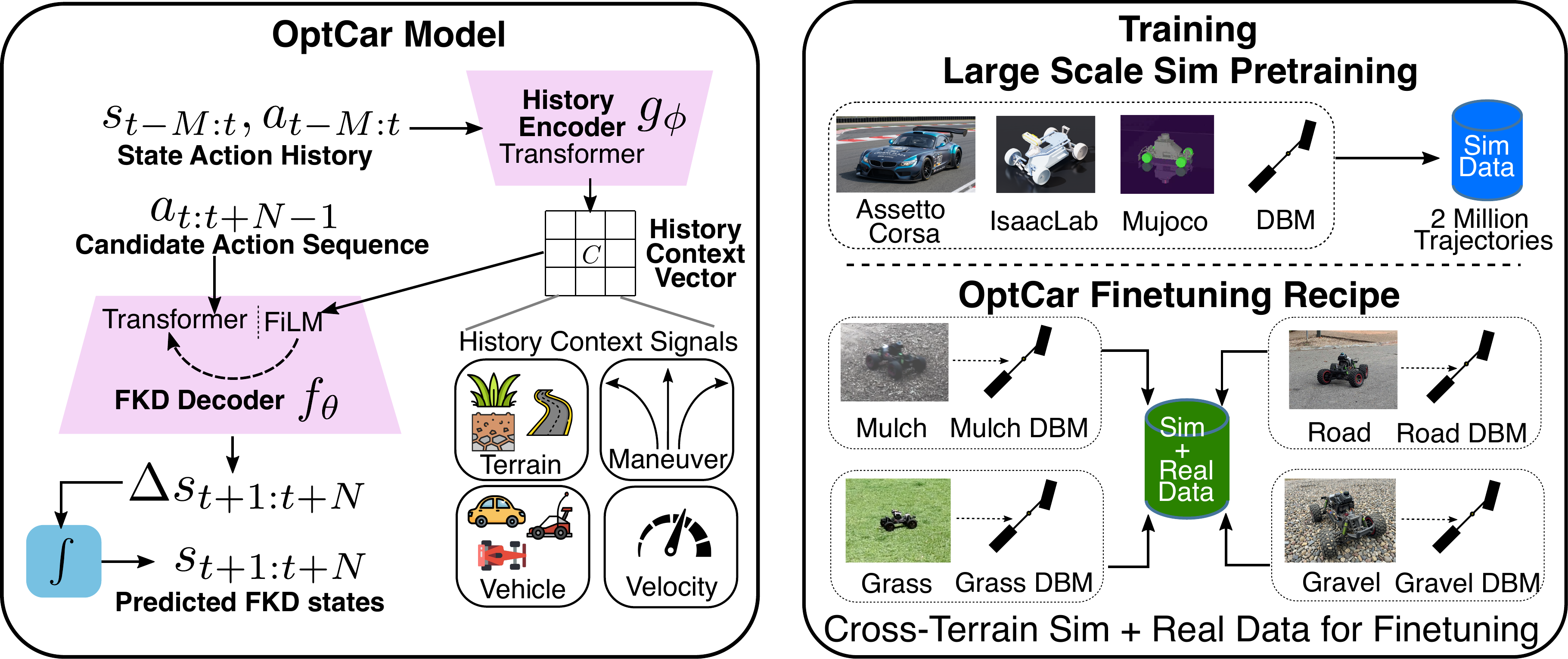}
    \caption{\textbf{\ourmethod{} overview.} \textit{Left:} Recent state-action history is compressed into a dynamics context vector $\mathbf{c}$ that conditions FKD rollouts for candidate future actions. \textit{Right:} \ourmethod{} is pretrained as a generalist FKD prior on simulation, then specialized to the target vehicle by fine-tuning on real data combined with targeted synthetic rollouts from per-terrain system identification. \vspace{-10pt}}
    \label{fig:architecture}
\end{figure}

\ourmethod{} specializes a pretrained FKD model to a target vehicle while adapting across terrains (Figure~\ref{fig:architecture}). The approach is built around two ideas. First, changing vehicle-terrain interaction is treated as a latent dynamics condition inferred online from recent motion; this inferred context then conditions all candidate future rollouts evaluated by MPC. Second, target-vehicle specialization is performed with limited real data augmented by targeted synthetic rollouts from terrain-specific system identification, expanding coverage of high-slip regimes that are difficult to collect safely.
\vspace{-10pt}

\subsection{History-Conditioned FKD Prediction}
\label{sec:optcar_arch}
\vspace{-10pt}

At high speed, the future motion of the vehicle is not determined by the measured state and commanded action alone. The same action sequence can produce different rollouts depending on latent factors such as terrain friction, slip regime, payload, and actuator response. These factors are not directly observed by the controller, but recent state-action transitions provide a local system-identification signal. \ourmethod{} therefore realizes the predictor of Eq.~\eqref{eq:fkd} in two stages: a history-to-context map $g_\phi$ and a context-conditioned rollout map $f_\theta$ whose predicted increments integrate to the future trajectory:
\begin{equation}
    \mathbf{c}_t = \underbrace{g_\phi\!\left(\mathbf{s}_{t-M:t},\; \mathbf{a}_{t-M:t}\right)}_{\text{history-to-context map}}; \quad
    \Delta\hat{\mathbf{s}}_{t+1:t+N} = \underbrace{f_\theta\!\left(\mathbf{s}_t,\; \mathbf{a}_{t:t+N-1},\; \mathbf{c}_t\right)}_{\text{context-conditioned rollout map}}.
    \label{eq:factorization}
\end{equation}
The context $\mathbf{c}_t \in \mathbb{R}^d$ is not supervised with terrain or regime labels; it is learned end-to-end as the information from recent motion that is useful for predicting future rollouts. This factorization creates a bottleneck between history processing and rollout prediction: all history-dependent information available to the rollout model must pass through $\mathbf{c}_t$. Rather than cross-attending to individual history tokens, \ourmethod{} conditions every block of the rollout map $f_\theta$ on $\mathbf{c}_t$:
\begin{equation}
    \mathbf{h}^{(l+1)} = \mathcal{B}^{(l)}_\theta\!\left(\mathbf{h}^{(l)};\, \mathbf{c}_t\right), \qquad l = 1,\dots,L,
    \label{eq:block_conditioning}
\end{equation}
where $\mathcal{B}^{(l)}_\theta$ is the $l$-th block of the rollout map $f_\theta$ and $\mathbf{h}^{(l)}$ its hidden representation, with conditioning realized by FiLM~\cite{perez2017filmvisualreasoninggeneral} in our implementation. Every candidate rollout for the sampling-based MPC is conditioned on the same dynamics context, and $\mathbf{c}_t$ is computed once per planning step and reused across all candidate action sequences.

The decoder predicts body-frame pose increments $\Delta\hat{\mathbf{s}}_{t+k} = (\Delta\hat{\mathbf{p}}_{t+k}, \Delta\hat{\mathbf{q}}_{t+k})$ rather than global poses. This local representation removes dependence on global translation and supports rollouts over non-planar terrain. Architecture and training details are summarized in Section~\ref{sec:impl}, with exact conditioning and integration equations in Appendix~\ref{app:impl}.
\vspace{-6pt}

\subsection{Pretraining as a Generalist FKD Prior}
\label{sec:pretraining}
\vspace{-6pt}
Before target-vehicle specialization, \ourmethod{} is pretrained on a large simulation corpus spanning diverse vehicles, terrains, and driving regimes. The purpose of pretraining is to learn a broad prior over action-conditioned vehicle motion and to learn how recent state-action history can identify the active dynamics. Both the history-to-context map $g_\phi$ and the rollout map $f_\theta$ are trained jointly from random initialization. The training target is the body-frame future increment sequence in Eq.~\eqref{eq:factorization}; corpus and loss details are summarized in Section~\ref{sec:impl} and Appendix~\ref{app:impl}.
\vspace{-6pt}
\subsection{Targeted Real-and-Synthetic Specialization}
\label{sec:finetuning}
\vspace{-6pt}
Fine-tuning must balance relevance and coverage. Real target-vehicle data are most relevant but sparse in the aggressive high-slip regimes that matter for MPC, while broad simulation randomization adds coverage at the cost of capacity spent on unrelated vehicles and terrains. \ourmethod{} instead uses each per-terrain real dataset twice, as direct training data and as the calibration signal for a terrain-specific synthetic data generator anchored to the same vehicle.

Let~$\mathcal{D}^e_{\text{real}}$ denote the real target-vehicle trajectories collected on terrain $e \in \mathcal{E}$. For each terrain, we identify the parameters $\psi_e$ of a dynamic bicycle model (DBM) $F_{\psi_e}$ by fitting one-step transitions:
\begin{equation}
    \psi_e^* = \arg\min_{\psi_e}
    \sum_{(\mathbf{s}_t,\mathbf{a}_t,\mathbf{s}_{t+1}) \in \mathcal{D}^e_{\text{real}}}
    \left\| F_{\psi_e}(\mathbf{s}_t,\mathbf{a}_t) - \mathbf{s}_{t+1} \right\|^2.
    \label{eq:sysid}
\end{equation}
These calibrated terrain-specific models are also used in the DBM-MPC baseline. The calibrated model $F_{\psi_e^*}$ is not deployed as \ourmethod{}'s rollout model. Instead, it is used offline to generate targeted synthetic rollouts $\mathcal{D}^e_{\text{synth}}$ under randomized action sequences that enter high-slip regions under-sampled by the real data while remaining tied to the identified vehicle-terrain dynamics. The final fine-tuning set is
\begin{equation}
    \mathcal{D}_{\text{FT}} =
    \bigcup_{e \in \mathcal{E}}
    \left(\mathcal{D}^e_{\text{real}} \cup \mathcal{D}^e_{\text{synth}}\right),
    \label{eq:ft_dataset}
\end{equation}
and the pretrained model is fine-tuned end-to-end on $\mathcal{D}_{\text{FT}}$ with the same pretraining prediction loss.
\vspace{-8pt}
\subsection{Deployment in Sampling-Based MPC}
\label{sec:deployment}
\vspace{-6pt}
The fine-tuned model is the rollout function inside an MPPI controller~\cite{MPPI_2015}. At each planning step the history window is encoded into $\mathbf{c}_t$, candidate actions are rolled out with $f_\theta(\cdot,\cdot,\mathbf{c}_t)$, and the first action of the cost-weighted average of the sampled action sequences is applied. Since adaptation occurs through the forward pass for $\mathbf{c}_t$, deployment needs no terrain classifier, mode switch, or online parameter update.
\vspace{-6pt}

\section{Implementation Details}
\label{sec:impl}
\vspace{-8pt}

\paragraph{Architecture.}
\ourmethod{} uses the transformer FKD backbone and simulation protocol of AnyCar~\cite{anycar}, with history processing factored as in Eq.~\eqref{eq:factorization}. Historical state deltas, actions, and current-state features are embedded by per-stream MLPs into a shared dimension $d=64$. Historical and candidate future actions share the same embedding MLP. The history encoder is a 2-layer bidirectional transformer that maps $M+1=250$ history entries at $\Delta t=0.02$~s to the context vector $\mathbf{c}_t$. The rollout decoder is a 2-block causal transformer with 4 attention heads and feed-forward dimension 256, conditioned by $\mathbf{c}_t$ through FiLM at each block, and predicts $N=50$ body-frame SE(3) increments. Appendix~\ref{app:impl} gives the conditioning equations and body-frame integration.
\vspace{-8pt}

\paragraph{Training.}
The hyperparameters for the shared FKD backbone and optimization follow the AnyCar protocol~\cite{anycar} and are kept identical for \ourmethod{} and the AnyCar baseline.
We pretrain on approximately 2 million simulated history--future windows from the AnyCar protocol across Assetto Corsa, MuJoCo, Isaac Lab, and randomized analytical dynamic bicycle models. The objective is $\mathcal{L}=1.0\,\mathcal{L}_{\text{pos}}+0.3\,\mathcal{L}_{\text{quat}}$ (Appendix~\ref{app:impl}); optimization uses AdamW with weight decay $10^{-2}$, learning rate $10^{-3}$, cosine annealing, linear warmup, 500 epochs, and batch size 4096.

Fine-tuning uses approximately 5 minutes of real driving per terrain on road, gravel, grass, and mulch, yielding about 60K real history+future windows. The real fine-tuning data includes 6~m/s maneuvers. Targeted simulation adds about 400K windows by rolling out each $F_{\psi_e^*}$ for 10~s at $\Delta t=0.02$~s under randomized B\'ezier velocity and random-walk steering commands within $[-2,8]$~m/s and $[-0.41,0.41]$~rad, with velocity noise $\sigma=0.2$. We reserve 10\% of the driving runs for validation and construct training and validation windows from separate runs. The combined real-and-synthetic fine-tuning dataset is randomly shuffled. Fine-tuning uses the same optimizer for 200 epochs at learning rate $10^{-3}$ and batch size 4096.
\vspace{-8pt}

\paragraph{Deployment.}
At each planning step, we find the point on the reference curve nearest to the current vehicle position and sample $H$ reference points ahead at equal time intervals $\Delta t=0.02$~s, advancing along the curve at $v_{\text{ref}}$. The reference point for lookahead step $k$ is therefore $v_{\text{ref}}k\Delta t$ along the curve from the nearest point and is shared by all candidate rollouts. MPPI samples 600 candidate action sequences per step and scores each rollout using
\begin{equation}
    C = \sum_{k=1}^{H} \gamma^k
    \left[
        w_{\text{long}} e_{\text{long},k}^2 + w_{\text{lat}} e_{\text{lat},k}^2
        + w_\psi \Delta\psi_k^2
        + w_v (v_k - v_{\text{ref}})^2
    \right],
    \label{eq:mppi_cost}
\end{equation}
where $H=N=50$, $\gamma=0.97$, $w_{\text{long}}=5$, $w_{\text{lat}}=75$, $w_\psi=1$, and $w_v=15$. Here, $e_{\text{long},k}$ and $e_{\text{lat},k}$ are longitudinal and lateral tracking errors, $\Delta\psi_k$ is heading error, $v_k$ is predicted vehicle speed, and $v_{\text{ref}}$ is commanded speed. The system is implemented in PyTorch and CuPy and runs onboard an NVIDIA Jetson AGX Orin. Both \ourmethod{} and AnyCar use FP16 TensorRT inference for onboard deployment. With 600 candidate rollouts and a 50-step prediction horizon, \ourmethod{} replans at 37~Hz, compared with 32~Hz for AnyCar.
\vspace{-10pt}

\section{Experiments}
\label{sec:experiments}
\vspace{-8pt}
We design experiments to answer three research questions:
\vspace{-8pt}
\begin{itemize}
    \setlength{\itemsep}{1pt}
    \setlength{\parskip}{1pt}
    \item[\textbf{RQ1.}] \textbf{Specialized generalist versus baselines:} Can \ourmethod{} match or exceed both generalist and specialist models in cross-terrain tracking and out-of-distribution dynamics settings?
    \item[\textbf{RQ2.}] \textbf{Targeted real+sim fine-tuning:} Does fine-tuning on limited real data combined with targeted synthetic rollouts improve closed-loop tracking over fine-tuning on just real data?
    \item[\textbf{RQ3.}] \textbf{History Conditioning:} Does FiLM-based history conditioning improve closed-loop tracking compared with context concatenation and cross-attention over history tokens? 
\end{itemize}
\vspace{-10pt}

\subsection{Experimental Setup}
\label{sec:exp_setup}

\begin{figure*}[t]
    \centering
    \settoheight{\platformht}{\includegraphics[width=0.49\textwidth]{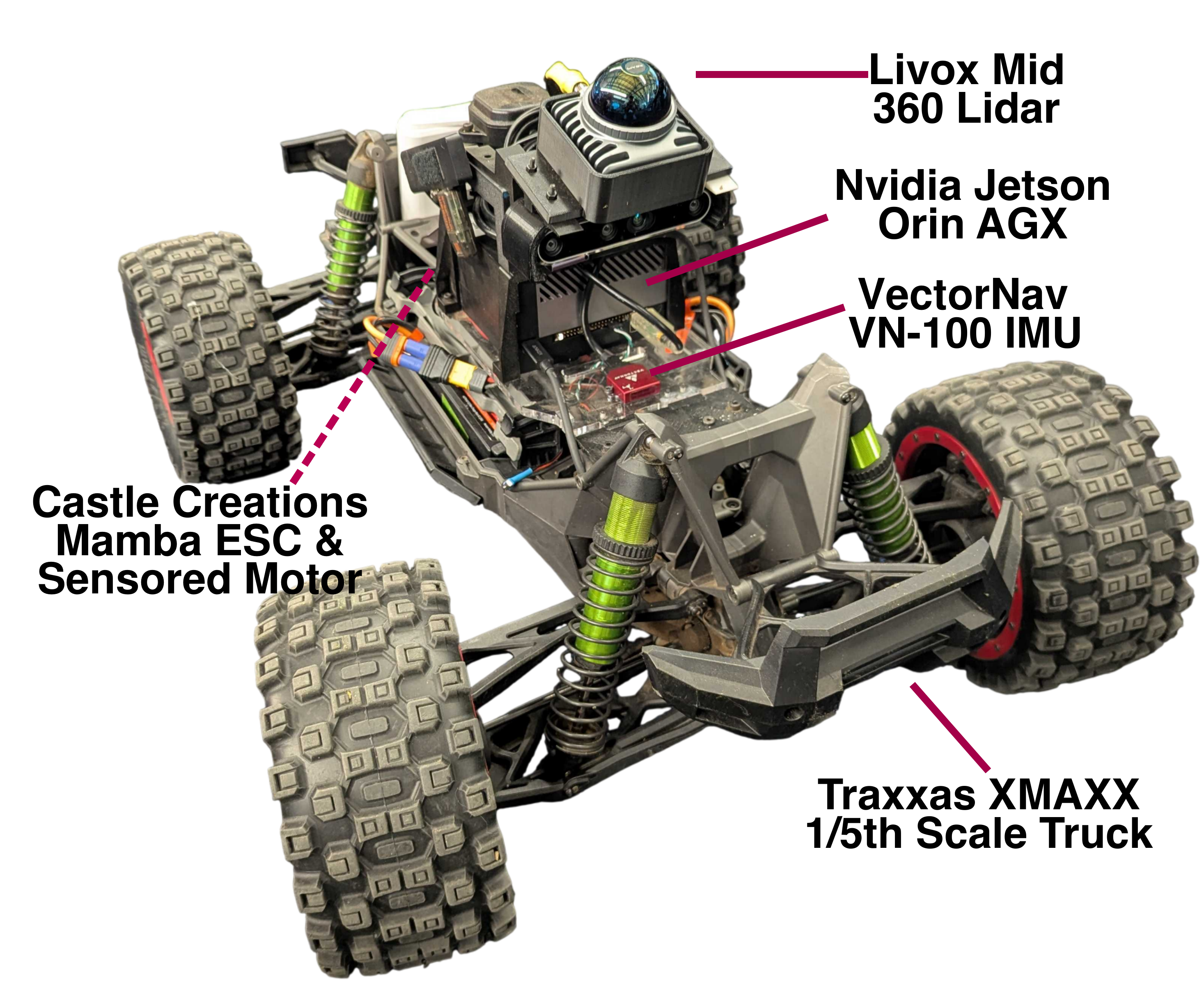}}%
    \setlength{\envimght}{0.5\platformht}%
    \addtolength{\envimght}{-13pt}%
    \begin{subfigure}[c]{0.49\textwidth}
        \centering
        \includegraphics[width=\textwidth]{images/platform.pdf}
        \caption{Platform components}
    \end{subfigure}
    \hfill
    \begin{minipage}[c]{0.49\textwidth}
        \centering
        \begin{subfigure}[t]{0.49\linewidth}
            \centering
            \includegraphics[width=\textwidth,height=\envimght,clip,keepaspectratio=false]{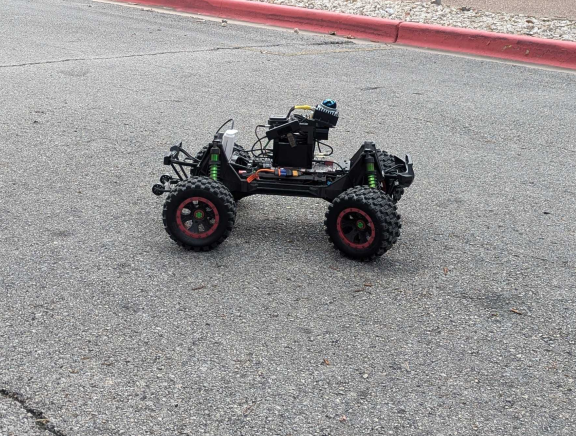}
            \caption{Road}
        \end{subfigure}
        \hfill
        \begin{subfigure}[t]{0.49\linewidth}
            \centering
            \includegraphics[width=\textwidth,height=\envimght,clip,keepaspectratio=false]{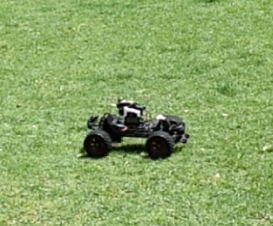}
            \caption{Wet Grass + Slope}
        \end{subfigure}

        \vspace{1ex}

        \begin{subfigure}[t]{0.49\linewidth}
            \centering
            \includegraphics[width=\textwidth,height=\envimght,clip,keepaspectratio=false]{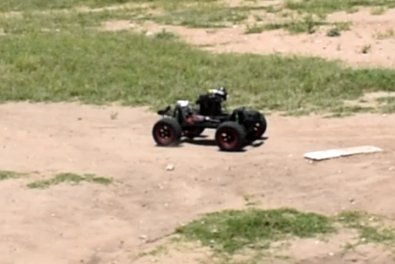}
            \caption{Vegetation + Dirt}
        \end{subfigure}
        \hfill
        \begin{subfigure}[t]{0.49\linewidth}
            \centering
            \includegraphics[width=\textwidth,height=\envimght,clip,keepaspectratio=false]{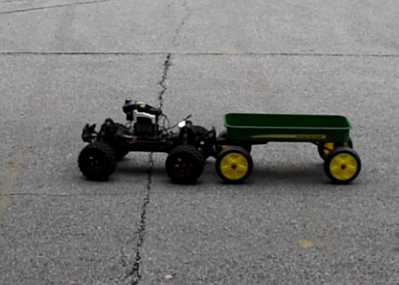}
            \caption{Cart Pulling}
        \end{subfigure}
    \end{minipage}
    \caption{\textbf{Experimental platform and test environments.} Platform sensor and compute stack, followed by the terrain and payload conditions used for figure-8 tracking. \vspace{-8pt}}
    \label{fig:exp_setup}
    \vspace{-10pt}
\end{figure*}

\paragraph{Platform setup:}
All experiments use the 1/5-scale AlphaTruck~\cite{rana2026alphatruckplatformhighspeedoffroad} off-road vehicle with onboard 3D LiDAR, IMU, and NVIDIA Jetson AGX compute (Figure~\ref{fig:exp_setup}). SuperOdometry~\cite{zhao2021super} provides localization, and a low-level PD controller maps commanded velocities to PWM signals.
\vspace{-4pt}

\paragraph{Reference trajectories and terrains:}
We evaluate closed-loop figure-8 tracking at three commanded speeds: easy ($v_{\text{ref}} = 2$~m/s), medium ($v_{\text{ref}} = 4$~m/s), and hard ($v_{\text{ref}} = 6$~m/s). Road uses a tighter 17~m figure-8, while wet grass with slope and vegetation + dirt use a 30~m figure-8 so that 6~m/s tracking remains feasible while preserving the high-slip modeling challenge. The trajectory parameterization is given in Appendix~\ref{app:reftraj}.
\vspace{-8pt}

\paragraph{Metrics:}
We report trajectory tracking error (TTE), the mean distance from the vehicle to the closest point on the reference trajectory, and velocity tracking error (VTE), the mean absolute deviation from the commanded speed. Means and standard deviations are computed over per-timestep tracking errors pooled across three runs.
\vspace{-8pt}

\paragraph{Baselines:}
We compare models along two axes: how history conditions the rollout decoder and the fine-tuning recipe. \textbf{AnyCar}~\cite{anycar} cross-attends to individual history tokens, while \textbf{\ourmethod{}} and \textbf{Concat} compress history into a single dynamics context token and inject it through FiLM or input concatenation, respectively. For fine-tuning, \textbf{FT-R} uses real target-vehicle data, while \textbf{FT-RS} additionally includes targeted synthetic rollouts from per-terrain system identification. We also compare against a terrain-calibrated dynamic bicycle model (\textbf{DBM}), \textbf{IKD FT-R} (Appendix~\ref{app:ikd}), and \textbf{\ourmethod{} Specialist}, trained only on 30 minutes of real road data. DBM and the IKD planner use road parameters for road and cart pulling, grass parameters for wet grass + slope, and dirt parameters for vegetation + dirt. \ourmethod{} is not fine-tuned on dirt, but is exposed to slippery terrains including mulch during fine-tuning and randomized dynamics during pretraining. AnyCar FT-RS, Concat FT-RS, and \ourmethod{} Specialist are evaluated only on road and wet grass + slope.
\vspace{-8pt}

\begin{table*}[t]
    \centering
    \small
    \begin{tabular}{ll cc cc cc}
        \toprule
        & & \multicolumn{2}{c}{\textbf{Easy (2 m/s)}} & \multicolumn{2}{c}{\textbf{Medium (4 m/s)}} & \multicolumn{2}{c}{\textbf{Hard (6 m/s)}} \\
        \cmidrule(lr){3-4} \cmidrule(lr){5-6} \cmidrule(lr){7-8}
        \textbf{Env.} & \textbf{Method} & TTE & VTE & TTE & VTE & TTE & VTE \\
        \midrule
        \multirow{8}{*}{Road}
        & DBM                 & \textbf{0.011} & \textbf{0.004}
                              & 0.034 & \textbf{0.019}
                              & 0.562 & 1.234 \\

        & IKD FT-R            & 0.016 & 0.032
                              & 0.067 & 0.189
                              & 0.412 & 1.849 \\

        & AnyCar FT-R         & 0.013 & 0.009
                              & 0.025 & 0.025
                              & 0.248 & 0.912 \\

        & AnyCar FT-RS        & 0.014 & 0.008
                              & 0.028 & 0.022
                              & 0.152 & 0.882 \\

        & Concat FT-RS        & 0.016 & 0.009
                              & 0.027 & 0.035
                              & 0.176 & 1.002 \\

        \arrayrulecolor{black!85}\cdashline{2-8}[0.6pt/1.4pt]\arrayrulecolor{black}
        & \ourmethod{} Specialist    & 0.012 & 0.007
                                     & \textbf{0.020} & 0.020
                                     & \textbf{0.093} & \textbf{0.711} \\

        & \ourmethod{} FT-R          & 0.018 & \textbf{0.004}
                                     & 0.034 & 0.027
                                     & 0.196 & 0.911 \\

        & \ourmethod{} FT-RS         & 0.016 & 0.013
                                     & 0.026 & 0.023
                                     & 0.106 & 0.901 \\

        \midrule
        \multirow{8}{*}{\shortstack[l]{Wet Grass\\+ Slope}}
        & DBM                        & 0.012 & 0.017
                                     & 0.156 & 0.044
                                     & 1.356 & 1.456 \\

        & IKD FT-R                   & 0.013 & 0.029
                                     & 0.062 & 0.255
                                     & 2.371 & 1.004 \\

        & AnyCar FT-R                & \textbf{0.011} & 0.025
                                     & 0.100 & 0.083
                                     & 1.082 & \textbf{0.892} \\

        & AnyCar FT-RS               & 0.014 & 0.022
                                     & 0.052 & 0.049
                                     & 0.771 & 0.922 \\

        & Concat FT-RS               & 0.015 & 0.026
                                     & 0.058 & 0.067
                                     & 0.785 & 1.083 \\

        \arrayrulecolor{black!85}\cdashline{2-8}[0.6pt/1.4pt]\arrayrulecolor{black}
        & \ourmethod{} Specialist    & 0.013 & \textbf{0.016}
                                     & 0.146 & 0.051
                                     & 1.149 & 1.056 \\

        & \ourmethod{} FT-R          & 0.013 & 0.027
                                     & \textbf{0.031} & 0.078
                                     & 0.751 & 0.933 \\

        & \ourmethod{} FT-RS         & 0.012 & 0.025
                                     & 0.044 & \textbf{0.011}
                                     & \textbf{0.625} & 0.940 \\

        \midrule
        \multirow{5}{*}{\shortstack[l]{Vegetation\\+ Dirt}}
        & DBM                        & 0.061 & 0.041
                                     & 0.276 & 0.312
                                     & 0.824 & 1.118 \\

        & IKD FT-R                   & 0.055 & 0.033
                                     & 0.232 & 0.452
                                     & 0.669 & 1.452 \\

        & AnyCar FT-R                & \textbf{0.054} & \textbf{0.019}
                                     & 0.207 & 0.199
                                     & 0.776 & 0.991 \\

        \arrayrulecolor{black!85}\cdashline{2-8}[0.6pt/1.4pt]\arrayrulecolor{black}
        & \ourmethod{} FT-R          & 0.055 & 0.020
                                     & 0.171 & 0.172
                                     & 0.519 & 0.981 \\

        & \ourmethod{} FT-RS         & 0.059 & 0.023
                                     & \textbf{0.157} & \textbf{0.072}
                                     & \textbf{0.348} & \textbf{0.937} \\

        \midrule
        \multirow{5}{*}{\shortstack[l]{Cart\\Pulling}}
        & DBM                        & 0.129 & 0.014
                                     & 0.212 & 0.197
                                     & 1.012 & 1.881 \\

        & IKD FT-R                   & 0.067 & 0.086
                                     & 0.244 & 0.102
                                     & 0.711 & 1.782 \\

        & AnyCar FT-R                & 0.042 & 0.031
                                     & 0.182 & 0.049
                                     & 0.489 & 1.338 \\

        \arrayrulecolor{black!85}\cdashline{2-8}[0.6pt/1.4pt]\arrayrulecolor{black}
        & \ourmethod{} FT-R          & 0.028 & 0.028
                                     & 0.174 & 0.062
                                     & 0.319 & 1.290 \\

        & \ourmethod{} FT-RS         & \textbf{0.011} & \textbf{0.014}
                                     & \textbf{0.085} & \textbf{0.044}
                                     & \textbf{0.229} & \textbf{1.023} \\

        \bottomrule

    \end{tabular}

    \caption{\textbf{Closed-loop figure-8 tracking.} TTE (m~$\downarrow$) and VTE (m/s~$\downarrow$) for \ourmethod{} and baselines across terrains, speeds, and the unseen cart-pulling payload. We compare history-conditioning designs (Concat, \ourmethod{}, and AnyCar), fine-tuning recipes (FT-R, FT-RS, and Specialist), and DBM and IKD baselines. Results with standard deviations are provided in Appendix~\ref{app:tracking_std}, Table~\ref{tab:fig8_std}.\vspace{-8pt}}
    \label{tab:fig8}
\end{table*}

\begin{figure*}[t]
    \centering
    \includegraphics[width=0.95\textwidth]{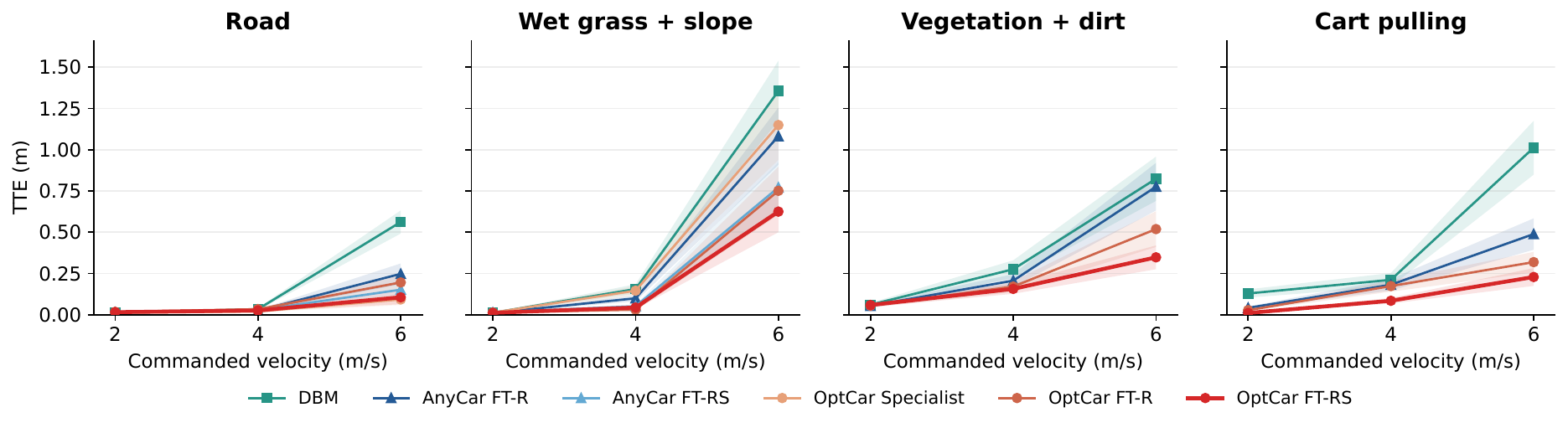}
    \caption{\textbf{Closed-loop tracking across speeds.} TTE (m~$\downarrow$) for \ourmethod{} and baselines across terrains and the unseen cart-pulling payload. Shaded bands indicate $\pm$1 standard deviation. Each panel shows the subset of these methods evaluated in that condition.}
    \label{fig:tte_velocity_main}
    \vspace{-15pt}
\end{figure*}

\subsection{RQ1: Baseline Comparisons}
\label{sec:exp_traj_tracking}

We evaluate \ourmethod{} against generalist, specialist, and analytical baselines in three settings: cross-terrain figure-8 tracking, an unseen cart-pulling payload, and transfer away from a road-only specialist model. Table~\ref{tab:fig8} summarizes the closed-loop tracking results, and Figure~\ref{fig:tte_velocity_main} shows how TTE varies with commanded velocity across environments.

\begin{figure*}[t]
    \centering
    \includegraphics[width=\textwidth]{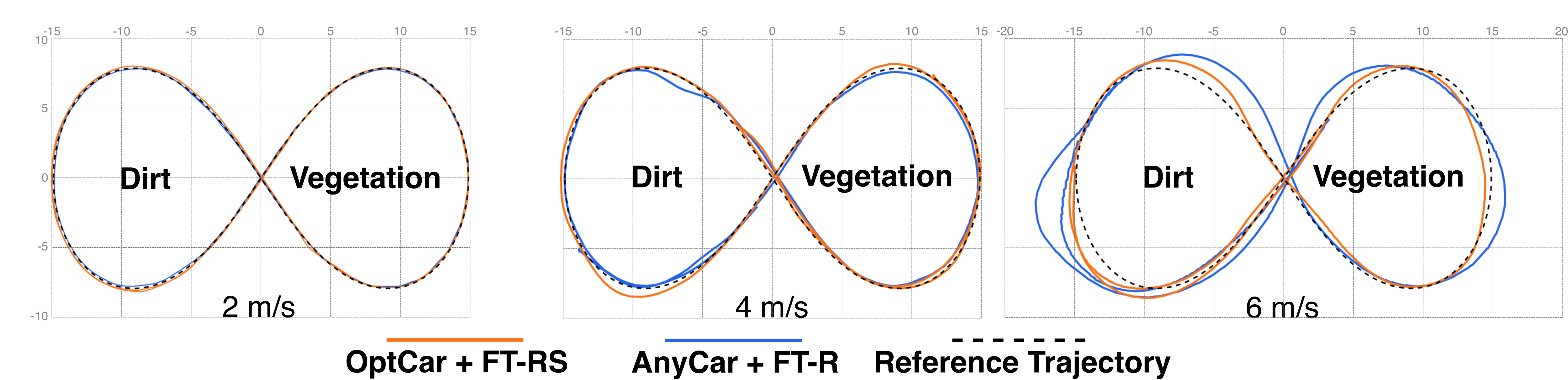}

    \caption{\textbf{\ourmethod{} FT-RS vs. AnyCar FT-R on vegetation + dirt.} Bird's-eye view of figure-8 tracking in the vegetation + dirt environment at three speeds. Slip is more pronounced in the dirt region, where \ourmethod{} FT-RS better maintains the reference trajectory. Grid axes are in meters. \vspace{-12pt}}
    \label{fig:fig8_rollouts}
    \vspace{-4 pt}
\end{figure*}

\subsubsection{Figure-8 Trajectory Tracking in Three Test Environments}
\vspace{-4pt}
The vehicle executes figure-8 trajectories on road, wet grass with slope, and vegetation + dirt at 2, 4, and 6~m/s. Each configuration is repeated three times for 15~s, and all models use identical MPPI parameters. Tracking accuracy is comparable across methods at low slip and diverges as speed and terrain-induced slip increase. At 2~m/s, all methods operate near the nominal regime and perform similarly. At 4~m/s, performance remains close on road, whose high traction keeps the dynamics predictable, while differences emerge on the lower-traction terrains. The separation is clearest at 6~m/s: \ourmethod{} FT-RS achieves lower trajectory tracking error than AnyCar FT-R on all three terrains, and on vegetation + dirt its error is less than half that of AnyCar FT-R (Figure~\ref{fig:fig8_rollouts}). IKD FT-R degrades at high speed, as it tracks states produced by a separate DBM-MPPI planner rather than scoring candidate actions with the same learned dynamics used for execution.

\vspace{-4 pt}
\subsubsection{Out-of-Distribution Cart Pulling}
\label{sec:exp_cart}
\begin{figure}[t]
    \centering
    \includegraphics[width=1.0\columnwidth]{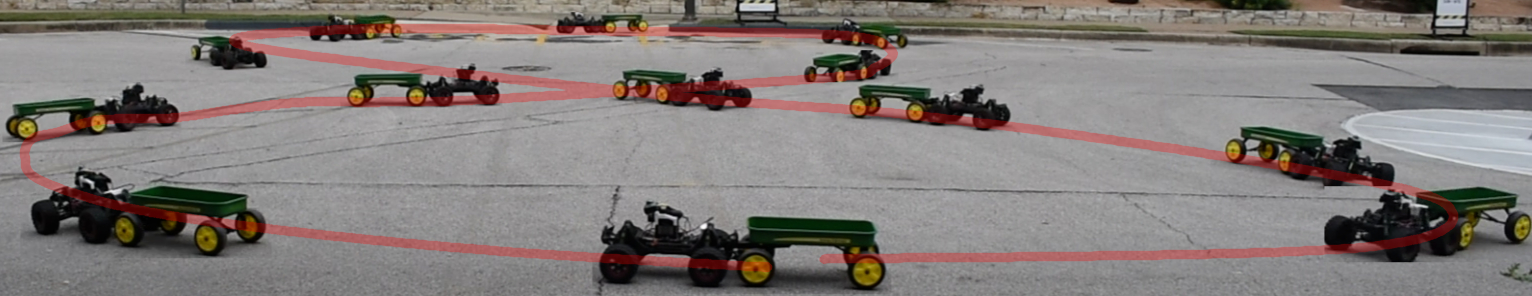}
    \caption{\textbf{Zero-shot cart pulling.} \ourmethod{} FT-RS tracking a figure-8 at 6~m/s with a trailing cart. No cart data is used during training. \vspace{-15pt}}
    \label{fig:cart}
    \vspace{-2pt}
\end{figure}

\vspace{-4pt}
The cart-pulling task evaluates whether adaptation extends beyond the terrains seen during fine-tuning. The trailing cart alters the vehicle's effective inertia and tire loading, and no cart data is used during training. \ourmethod{} FT-RS attains the lowest tracking error across all speeds, with the largest margin at 6~m/s (Table~\ref{tab:fig8}, Figure~\ref{fig:cart}). This demonstrates robustness to an unseen payload shift without cart-specific training data.
\vspace{-10 pt}

\subsubsection{Comparison Against a Specialist Model}
\label{sec:exp_specialist}
\vspace{-4pt}
The road specialist provides a single-terrain reference: it uses the same \ourmethod{} architecture but is fine-tuned only on 30 minutes of real road data. On road, the specialist attains the best 6~m/s tracking, and \ourmethod{} FT-RS remains close despite using only 5 minutes of real data per terrain. On wet grass with slope, the specialist's accuracy degrades and is well behind \ourmethod{} FT-RS. This illustrates the RQ1 tradeoff: the road specialist is accurate in-distribution but transfers poorly to wet grass + slope, whereas multi-terrain specialization remains accurate in both conditions.
\vspace{-7pt}
\subsection{RQ2: Targeted Real+Sim Fine-Tuning}
\label{sec:exp_finetuning}
\vspace{-4pt}
To evaluate the targeted real-and-synthetic fine-tuning recipe, we compare \ourmethod{} FT-R against \ourmethod{} FT-RS in Table~\ref{tab:fig8}. Both use the same history-conditioned architecture and real target-vehicle data; \ourmethod{} FT-RS additionally trains on synthetic rollouts generated from terrain-specific system identification. The real-only model remains competitive in moderate regimes, but the real+sim model is consistently stronger at 6~m/s, where aggressive slip dominates closed-loop error. \ourmethod{} FT-RS reduces trajectory tracking error at 6~m/s by 17\%--46\% relative to \ourmethod{} FT-R across all four conditions. On road and wet grass + slope, real-and-synthetic fine-tuning also improves AnyCar FT-RS over AnyCar FT-R at 6~m/s, reducing trajectory tracking error by 29\%--39\%. This supports the fine-tuning recipe: real data anchors the model to the target vehicle, while targeted synthetic data fills high-slip state-action regions that are difficult to collect safely.

\vspace{-4pt}
\subsection{RQ3: History Conditioning}
\label{sec:exp_context}
\vspace{-4pt}

To isolate the effect of the history-conditioning design, we compare Concat FT-RS, AnyCar FT-RS, and \ourmethod{} FT-RS on road and wet grass + slope in Table~\ref{tab:fig8}. All three are fine-tuned on the same real and targeted synthetic data; \ourmethod{} FT-RS uses a history context encoder and FiLM-conditioned decoder, whereas Concat supplies context as an additional input and AnyCar cross-attends to history tokens. Among the three conditioning designs, FiLM achieves the lowest trajectory tracking error at 4 and 6~m/s on both reported terrains. The gains are largest at 6~m/s, where \ourmethod{} FT-RS reduces trajectory tracking error relative to AnyCar FT-RS by 30\% on road and 19\% on wet grass with slope. A more extensive analysis of the learned context vector, projected with PCA and LDA, examines its variation across simulated vehicles, speeds, and turn directions (Appendix~\ref{app:ctx}).
\vspace{-8pt}

\section{Conclusion}
\vspace{-8pt}
We present \ourmethod{}, a recipe that specializes a generalist FKD foundation model to a vehicle for real-time MPC that stays accurate as the terrain changes. Our experiments yield three lessons. First, a single model can be both specialized to one vehicle and robust across terrains. Starting from a generalist prior and conditioning it on recent motion makes the model accurate on the target vehicle, while a specialist trained on a single terrain fails once the terrain changes. Second, compressing recent history into a single dynamics context token and conditioning the rollout decoder through FiLM enables adaptation to changing vehicle-terrain dynamics, improving high-speed tracking without online optimization. Third, targeted synthetic data from per-terrain system identification helps address the coverage-relevance trade-off of fine-tuning. Together, with only 5 minutes of real data per terrain, \ourmethod{} roughly halves 6~m/s tracking error relative to AnyCar FT-R and is competitive with a 30-minute single-terrain specialist on its home terrain while remaining accurate elsewhere. More broadly, FiLM-based history conditioning paired with targeted system-identification data helps in deploying foundation dynamics models on a specific robot without collecting large per-terrain datasets.

\vspace{-8pt}
\paragraph{Limitations and future work.}
First, \ourmethod{} adapts only through its feedforward context vector and never updates its weights online, so it assumes the encountered dynamics lie within the pretraining and fine-tuning distribution. Genuinely novel dynamics would instead call for online model adaptation, as in meta-learning that updates the dynamics model on the fly~\cite{levy2025metalearningonlinedynamicsmodel}, and combining fast feedforward context with slower online weight updates is a promising middle ground. Second, real-time MPPI tightly constrains the rollout model. Even at about 300K parameters, it affords only a short $N=50$ horizon, because lengthening it slows each rollout and reduces the action samples evaluated per step. This myopic horizon motivates a scaling-law study of the trade-off between model capacity, rollout horizon, and sample budget. Third, \ourmethod{} adapts from proprioceptive history alone, and adding exteroceptive inputs such as RGB, geometric terrain (point clouds or elevation maps), or satellite-derived costmaps~\cite{rana2026overseec} could supply richer context and anticipate terrain changes before they appear in recent motion. Finally, planar DBM rollouts omit vertical motion, roll/pitch dynamics, and slope-dependent gravity, limiting synthetic coverage on steep or uneven terrain.
\vspace{-6pt}

\acknowledgments{
This work has taken place in the Autonomous Mobile Robotics Laboratory (AMRL) at UT Austin. AMRL research is supported in part by the National Science Foundation (OIA-2535195, CAREER-2046955, OIA-2219236, DGE-2125858, CCF-2319471). Any opinions, findings, and conclusions expressed in this material are those of the authors and do not necessarily reflect the views of the sponsors.
Joydeep Biswas holds a position as Visiting Professor with NVIDIA and receives financial compensation for this work. The terms of this arrangement have been reviewed and approved by the University of Texas at Austin in accordance with its policy on objectivity in research. \\
We thank Adam Uccello for his contributions to building the experimental platform.
}

\begingroup
\setlength{\bibsep}{6pt plus 1pt minus 1pt}
\bibliography{references}  
\endgroup

\newpage
\appendix
\section{Architecture and Training Details}
\label{app:impl}

\paragraph{Context extraction and FiLM conditioning.}
The history spans indices $t-M:t$, including the current step. Each historical state feature contains a position delta and relative quaternion expressed with respect to the preceding pose, together with linear and angular velocities in the instantaneous body frame (13D total). Actions are 2D, and current-state features comprise world-frame orientation and linear/angular velocities (10D), excluding absolute position. Separate MLPs embed these streams into $d=64$ features; the historical state and action token sequences and the single current-state token are concatenated along the sequence dimension, with learned positional encodings added to form $\mathbf{M}_t$. A learnable summary token $\texttt{[CTX]}$ is prepended before the bidirectional transformer encoder $g_\phi$. Candidate future actions use the same action MLP and, with future positional encodings, form the decoder input. Context extraction is given by
\begin{equation}
    \left[\,\mathbf{c}_t;\; \tilde{\mathbf{M}}_t\,\right] =
    \operatorname{Enc}_\phi\!\left(\left[\,\texttt{[CTX]};\; \mathbf{M}_t\,\right]\right).
    \label{eq:ctx}
\end{equation}
Only the summary output $\mathbf{c}_t$ is passed to the rollout decoder; the refined per-step tokens $\tilde{\mathbf{M}}_t$ are discarded. Each decoder block applies Feature-wise Linear Modulation at two sites:
\begin{equation}
    \tilde{\mathbf{h}}^{(l)}_j =
    \boldsymbol{\gamma}^{(l)}_j(\mathbf{c}_t) \odot \mathbf{h}^{(l)}_j
    + \boldsymbol{\beta}^{(l)}_j(\mathbf{c}_t),
    \qquad j \in \{\mathrm{attn},\mathrm{ffn}\},
    \label{eq:film_post}
\end{equation}
where $\mathbf{h}^{(l)}_j$ denotes decoder features within block $l$ after self-attention ($j=\mathrm{attn}$) or after the feed-forward sublayer at the block output ($j=\mathrm{ffn}$). The two sites use separate learned affine projections $\boldsymbol{\gamma}^{(l)}_j, \boldsymbol{\beta}^{(l)}_j : \mathbb{R}^d \to \mathbb{R}^d$, initialized so that $\boldsymbol{\gamma}^{(l)}_j \approx \mathbf{1}$ and $\boldsymbol{\beta}^{(l)}_j \approx \mathbf{0}$.

\paragraph{Body-frame rollout integration.}
The decoder head outputs body-frame pose increments $\Delta\hat{\mathbf{s}}_{t+k} = (\Delta\hat{\mathbf{p}}_{t+k}, \Delta\hat{\mathbf{q}}_{t+k})$. The rotational output $\Delta\hat{\mathbf{q}}_{t+k}$ is $L_2$-normalized to a unit quaternion before rollout integration and geodesic loss evaluation. The rollout used by MPC is recovered by integrating these increments from the current pose, taken as the local origin $(\hat{\mathbf{p}}_t, \hat{\mathbf{q}}_t) = (\mathbf{0}, \mathbf{q}_{\mathrm{id}})$:
\begin{equation}
    \hat{\mathbf{q}}_{t+k} = \hat{\mathbf{q}}_{t+k-1} \otimes \Delta\hat{\mathbf{q}}_{t+k}, \qquad
    \hat{\mathbf{p}}_{t+k} = \hat{\mathbf{p}}_{t+k-1} + \mathbf{R}\!\left(\hat{\mathbf{q}}_{t+k-1}\right) \Delta\hat{\mathbf{p}}_{t+k},
    \label{eq:euler}
\end{equation}
for $k = 1, \dots, N$, where $\mathbf{R}(\cdot)$ is the rotation matrix of a unit quaternion, $\otimes$ denotes quaternion multiplication, and $\mathbf{q}_{\mathrm{id}}$ is the identity quaternion. Linear and angular velocities are recovered from the predicted pose sequence using finite differences with the rollout timestep $\Delta t=0.02$~s; the resulting predicted velocity is used in the MPC cost.

\paragraph{Training loss.}
The pretraining and fine-tuning loss combines position MSE with a geodesic quaternion loss:
\begin{equation}
    \mathcal{L} = \lambda_p \sum_{k=1}^{N} \left\| \Delta\hat{\mathbf{p}}_{t+k} - \Delta\mathbf{p}_{t+k} \right\|^2 + \lambda_q \sum_{k=1}^{N} d_{\text{geo}}\!\left(\Delta\hat{\mathbf{q}}_{t+k},\, \Delta\mathbf{q}_{t+k}\right),
    \label{eq:loss}
\end{equation}
with $\lambda_p=1.0$ and $\lambda_q=0.3$. The geodesic quaternion distance is
\begin{equation}
    d_{\text{geo}}(\mathbf{q}_1, \mathbf{q}_2) = 2\, \operatorname{arctan2}\!\left(\left\|\operatorname{Im}(\mathbf{q}_1^{-1} \otimes \mathbf{q}_2)\right\|,\; \left|\operatorname{Re}(\mathbf{q}_1^{-1} \otimes \mathbf{q}_2)\right|\right),
    \label{eq:geodesic}
\end{equation}
which is invariant to the double-cover ambiguity $\mathbf{q} \sim -\mathbf{q}$.

\section{Reference Trajectory}
\label{app:reftraj}

The figure-8 reference trajectory is parameterized by $\theta \in [-\pi,\pi]$ as
\begin{equation}
    x(\theta) = \frac{A\cos\theta}{1+\sin^2\theta},
    \qquad
    y(\theta) = \frac{wA\sin\theta\cos\theta}{1+\sin^2\theta},
    \label{eq:fig8_ref}
\end{equation}
with width factor $w=1.5$ and heading given by the tangent direction $\psi(\theta)=\operatorname{atan2}(dy/d\theta, dx/d\theta)$. We use amplitude $A=8.3$ for the road trajectory, giving an approximately 17~m horizontal span, and $A=15$ for the lower-traction terrain trajectories, giving a 30~m horizontal span.

\section{Dynamic Bicycle Model with Wheel-Velocity Actuation}
\label{app:dbm}

Each per-terrain simulator $F_{\psi_e}$ in Section~\ref{sec:finetuning} is a modified dynamic bicycle model (DBM) with driven wheels, in which throttle is a desired wheel-edge linear velocity command $d_w$ with a first-order wheel-speed state $v_w$~\cite{rajamani2011vehicle}. The model is planar, with yaw-only rotation represented internally and exported as a quaternion. Table~\ref{tab:dbm-params} lists the parameters; those marked ID form the per-terrain vector $\psi_e$ identified by nonlinear least squares on real one-step transitions (Eq.~\ref{eq:sysid}).

\begin{table}[h]
\centering
\small
\setlength{\tabcolsep}{6pt}
\begin{tabular}{llc}
\toprule
\textbf{Symbol} & \textbf{Description} & \textbf{Src.} \\
\midrule
$m$ & Vehicle mass & Meas. \\
$I$ & Yaw inertia ($\frac{1}{12}m(l^2+b^2)$) & Approx. \\
$l_f,\,l_r$ & COM to front/rear axle distance & Assumed \\
$\psi_{\max}$ & Maximum steering angle & Limit \\
$v_{w,\min},\,v_{w,\max}$ & Min/max wheel linear velocity & Limit \\
$T_w$ & Wheel-speed time constant & ID \\
$C_a$ & Drive force constant & ID \\
$C_{t,f},\,C_{t,r}$ & Front/rear tire cornering stiffness & ID \\
$C_r$ & Rolling resistance coefficient & ID \\
$C_d$ & Quadratic drag coefficient & ID \\
\bottomrule
\end{tabular}
\caption{Dynamic bicycle model parameters. In the inertia approximation, $l$ and $b$ denote vehicle length and width, respectively, distinct from the COM-to-axle distances $l_f$ and $l_r$. \textbf{Src.} denotes the parameter source: \emph{Meas.} = directly measured from the platform; \emph{Approx.} = approximated from vehicle geometry; \emph{Assumed} = assumed by model symmetry; \emph{Limit} = platform-imposed limit; \emph{ID} = identified by least-squares system identification on real-world data.}
\label{tab:dbm-params}
\end{table}

\paragraph{State and inputs.}
The DBM evolves the planar state
\begin{equation}
x = \begin{bmatrix} p_x & p_y & \theta & v_{\mathrm{long}} & v_{\mathrm{lat}} & r & v_w \end{bmatrix}^\top,
\end{equation}
where $(p_x,p_y)$ is planar position, $\theta$ is yaw, $(v_{\mathrm{long}},v_{\mathrm{lat}})$ are body-frame velocities, $r$ is the yaw rate, and $v_w$ is the wheel-edge linear velocity. The control input is $u = [\,d_w\;\; \psi\,]^\top$, clamped to physical limits,
\begin{equation}
\tilde{d}_w = \mathrm{clip}(d_w, v_{w,\min}, v_{w,\max}), \qquad \tilde{\psi} = \mathrm{clip}(\psi, -\psi_{\max}, \psi_{\max}).
\end{equation}
For compatibility with the learned FKD model, the planar state is exported in the same 13D pose-velocity format used throughout the paper, with the out-of-plane components set to zero and yaw encoded as a quaternion.

\paragraph{Dynamics.}
Planar kinematics are
\begin{align}
\dot{p}_x &= v_{\mathrm{long}}\cos\theta - v_{\mathrm{lat}}\sin\theta, \\
\dot{p}_y &= v_{\mathrm{long}}\sin\theta + v_{\mathrm{lat}}\cos\theta, \\
\dot{\theta} &= r .
\end{align}
Front and rear slip angles are
\begin{align}
\alpha_f &= \tilde{\psi} - \mathrm{atan2}(v_{\mathrm{lat}} + r l_f,\; v_{\mathrm{long}}), \\
\alpha_r &= -\mathrm{atan2}(v_{\mathrm{lat}} - r l_r,\; |v_{\mathrm{long}}|) .
\end{align}
Drive, tire, and resistive forces are
\begin{align}
F_f^{\mathrm{drive}} = F_r^{\mathrm{drive}} &= \tfrac{1}{2} C_a (v_w - v_{\mathrm{long}}), \\
F_f^{\mathrm{tire}} &= C_{t,f}\,\alpha_f, \qquad F_r^{\mathrm{tire}} = C_{t,r}\,\alpha_r, \\
F_{\mathrm{roll}} &= C_r\,\mathrm{sign}(v_{\mathrm{long}}), \qquad F_{\mathrm{drag}} = C_d\,v_{\mathrm{long}}|v_{\mathrm{long}}| .
\end{align}
The net body-frame forces and yaw moment are
\begin{align}
F_{\mathrm{long}} &= F_f^{\mathrm{drive}}\cos\tilde{\psi} + F_r^{\mathrm{drive}} - F_f^{\mathrm{tire}}\sin\tilde{\psi} - F_{\mathrm{roll}} - F_{\mathrm{drag}}, \\
F_{\mathrm{lat}} &= F_f^{\mathrm{drive}}\sin\tilde{\psi} + F_f^{\mathrm{tire}}\cos\tilde{\psi} + F_r^{\mathrm{tire}}, \\
\tau_z &= F_f^{\mathrm{drive}} l_f \sin\tilde{\psi} + F_f^{\mathrm{tire}} l_f \cos\tilde{\psi} - F_r^{\mathrm{tire}} l_r,
\end{align}
yielding the body-frame and wheel-speed dynamics
\begin{align}
\dot{v}_{\mathrm{long}} &= \tfrac{1}{m}F_{\mathrm{long}} + v_{\mathrm{lat}} r, &
\dot{v}_{\mathrm{lat}} &= \tfrac{1}{m}F_{\mathrm{lat}} - v_{\mathrm{long}} r, \\
\dot{r} &= \tfrac{1}{I}\tau_z, &
\dot{v}_w &= \frac{\tilde{d}_w - v_w}{T_w} .
\end{align}
The continuous-time system is integrated with RK4 to form the discrete transition $F_{\psi_e}$ used for system identification and synthetic rollout generation.

\begin{figure*}[t]
    \centering
    \includegraphics[width=1.0\textwidth]{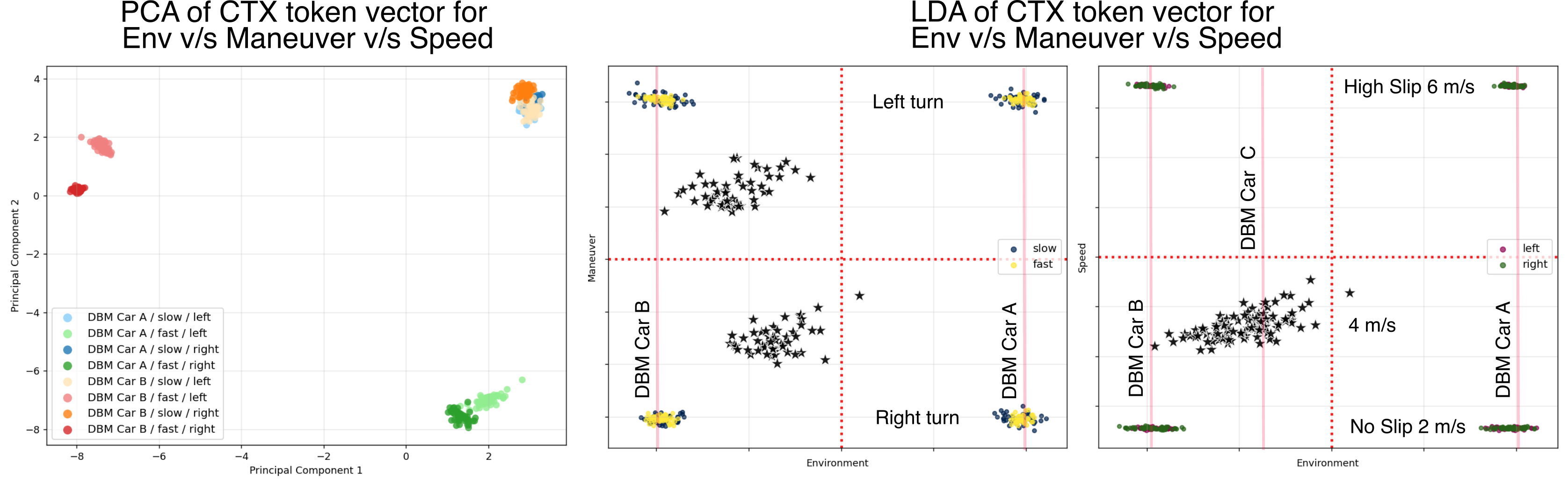}
    \caption{\textbf{Context-vector representation analysis.} Context vectors $\mathbf{c}$ extracted during simulated closed-loop operation are projected to two dimensions. \textit{Left:} PCA gives an unsupervised view of the embedding and separates the dominant dynamics regimes across vehicles, speeds, and turning directions. \textit{Middle:} LDA using maneuver labels separates left- and right-turn histories while preserving vehicle-specific structure along the environment axis. \textit{Right:} LDA using speed labels separates low-slip 2~m/s, moderate 4~m/s, and high-slip 6~m/s regimes, showing that the learned context encodes speed-dependent slip effects.}
    \label{fig:pca}
\end{figure*}

\section{Inverse Kinodynamic (IKD) Baseline}
\label{app:ikd}

The \textbf{IKD FT-R} baseline mirrors \ourmethod{}'s history-conditioned transformer, including the history encoder, the $\texttt{[CTX]}$ summary token, and FiLM conditioning, but reverses the input-output direction of prediction. Instead of predicting future states from a candidate action sequence, IKD takes a sequence of desired future states $\mathbf{s}^{\text{des}}_{t+1:t+N}$ and predicts the actions that realize them,
\begin{equation}
    \hat{\mathbf{a}}_{t:t+N-1} = h_\xi\!\left(\mathbf{s}_{t-M:t},\; \mathbf{a}_{t-M:t},\; \mathbf{s}^{\text{des}}_{t+1:t+N}\right),
    \label{eq:ikd}
\end{equation}
where the decoder consumes the candidate future-state stream and emits per-step throttle and steering commands. The desired future states are not free variables; they are produced by a separate planner that rolls out the per-terrain DBM $F_{\psi_e^*}$ (Appendix~\ref{app:dbm}) under MPPI to obtain a reference trajectory, which IKD then inverts into commands. Because execution depends on the planner's analytical model rather than on scoring action rollouts with the deployed learned dynamics, tracking accuracy is bounded by the validity of the DBM, which degrades in the high-slip regime and explains the sharp error growth of IKD FT-R at 6~m/s in Table~\ref{tab:fig8}.

\section{Context-Vector Representation Analysis}
\label{app:ctx}

We analyze the learned context vectors $\mathbf{c}_t$ extracted during simulated closed-loop operation across vehicles, speeds, and turning maneuvers by projecting them with PCA and LDA (Figure~\ref{fig:pca}). In the unsupervised PCA projection, contexts cluster by dominant dynamics regime, including vehicle, speed, and turning direction. LDA projections using maneuver and speed labels show that these factors are also linearly recoverable from the embedding. Thus the context vector is not merely a trajectory identifier; it organizes recent history around factors that affect rollout prediction for MPC.

\section{Closed-Loop Tracking Results with Standard Deviations}
\label{app:tracking_std}

Table~\ref{tab:fig8_std} reports means and standard deviations of per-timestep tracking errors pooled across three runs for all methods.

\begin{table}[H]
    \centering
    \small
    \begin{adjustbox}{max width=\textwidth}
    \begin{tabular}{ll cc cc cc}
        \toprule
        & & \multicolumn{2}{c}{\textbf{Easy (2 m/s)}} & \multicolumn{2}{c}{\textbf{Medium (4 m/s)}} & \multicolumn{2}{c}{\textbf{Hard (6 m/s)}} \\
        \cmidrule(lr){3-4} \cmidrule(lr){5-6} \cmidrule(lr){7-8}
        \textbf{Env.} & \textbf{Method} & TTE & VTE & TTE & VTE & TTE & VTE \\
        \midrule
        \multirow{8}{*}{Road}
        & DBM                 & \textbf{0.011} $\pm$ 0.003 & \textbf{0.004} $\pm$ 0.001
                              & 0.034 $\pm$ 0.007 & \textbf{0.019} $\pm$ 0.004
                              & 0.562 $\pm$ 0.070 & 1.234 $\pm$ 0.281 \\

        & IKD FT-R            & 0.016 $\pm$ 0.003 & 0.032 $\pm$ 0.007
                              & 0.067 $\pm$ 0.014 & 0.189 $\pm$ 0.040
                              & 0.412 $\pm$ 0.052 & 1.849 $\pm$ 0.423 \\

        & AnyCar FT-R         & 0.013 $\pm$ 0.002 & 0.009 $\pm$ 0.002
                              & 0.025 $\pm$ 0.005 & 0.025 $\pm$ 0.005
                              & 0.248 $\pm$ 0.063 & 0.912 $\pm$ 0.197 \\

        & AnyCar FT-RS        & 0.014 $\pm$ 0.003 & 0.008 $\pm$ 0.002
                              & 0.028 $\pm$ 0.006 & 0.022 $\pm$ 0.005
                              & 0.152 $\pm$ 0.041 & 0.882 $\pm$ 0.192 \\

        & Concat FT-RS        & 0.016 $\pm$ 0.003 & 0.009 $\pm$ 0.002
                              & 0.027 $\pm$ 0.005 & 0.035 $\pm$ 0.008
                              & 0.176 $\pm$ 0.064 & 1.002 $\pm$ 0.230 \\

        \arrayrulecolor{black!85}\cdashline{2-8}[0.6pt/1.4pt]\arrayrulecolor{black}
        & \ourmethod{} Specialist
                              & 0.012 $\pm$ 0.002 & 0.007 $\pm$ 0.002
                              & \textbf{0.020} $\pm$ 0.004 & 0.020 $\pm$ 0.004
                              & \textbf{0.093} $\pm$ 0.034 & \textbf{0.711} $\pm$ 0.161 \\

        & \ourmethod{} FT-R   & 0.018 $\pm$ 0.004 & \textbf{0.004} $\pm$ 0.001
                              & 0.034 $\pm$ 0.007 & 0.027 $\pm$ 0.006
                              & 0.196 $\pm$ 0.041 & 0.911 $\pm$ 0.209 \\

        & \ourmethod{} FT-RS  & 0.016 $\pm$ 0.003 & 0.013 $\pm$ 0.003
                              & 0.026 $\pm$ 0.005 & 0.023 $\pm$ 0.005
                              & 0.106 $\pm$ 0.042 & 0.901 $\pm$ 0.203 \\

        \midrule
        \multirow{8}{*}{\shortstack[l]{Wet Grass\\+ Slope}}
        & DBM                 & 0.012 $\pm$ 0.002 & 0.017 $\pm$ 0.004
                              & 0.156 $\pm$ 0.029 & 0.044 $\pm$ 0.010
                              & 1.356 $\pm$ 0.182 & 1.456 $\pm$ 0.330 \\

        & IKD FT-R            & 0.013 $\pm$ 0.003 & 0.029 $\pm$ 0.006
                              & 0.062 $\pm$ 0.012 & 0.255 $\pm$ 0.055
                              & 2.371 $\pm$ 0.421 & 1.004 $\pm$ 0.234 \\

        & AnyCar FT-R         & \textbf{0.011} $\pm$ 0.002 & 0.025 $\pm$ 0.005
                              & 0.100 $\pm$ 0.019 & 0.083 $\pm$ 0.018
                              & 1.082 $\pm$ 0.173 & \textbf{0.892} $\pm$ 0.202 \\

        & AnyCar FT-RS        & 0.014 $\pm$ 0.003 & 0.022 $\pm$ 0.005
                              & 0.052 $\pm$ 0.010 & 0.049 $\pm$ 0.011
                              & 0.771 $\pm$ 0.151 & 0.922 $\pm$ 0.215 \\

        & Concat FT-RS        & 0.015 $\pm$ 0.003 & 0.026 $\pm$ 0.006
                              & 0.058 $\pm$ 0.011 & 0.067 $\pm$ 0.015
                              & 0.785 $\pm$ 0.179 & 1.083 $\pm$ 0.250 \\

        \arrayrulecolor{black!85}\cdashline{2-8}[0.6pt/1.4pt]\arrayrulecolor{black}
        & \ourmethod{} Specialist
                              & 0.013 $\pm$ 0.003 & \textbf{0.016} $\pm$ 0.004
                              & 0.146 $\pm$ 0.028 & 0.051 $\pm$ 0.011
                              & 1.149 $\pm$ 0.211 & 1.056 $\pm$ 0.240 \\

        & \ourmethod{} FT-R   & 0.013 $\pm$ 0.003 & 0.027 $\pm$ 0.006
                              & \textbf{0.031} $\pm$ 0.006 & 0.078 $\pm$ 0.017
                              & 0.751 $\pm$ 0.146 & 0.933 $\pm$ 0.207 \\

        & \ourmethod{} FT-RS  & 0.012 $\pm$ 0.002 & 0.025 $\pm$ 0.005
                              & 0.044 $\pm$ 0.009 & \textbf{0.011} $\pm$ 0.003
                              & \textbf{0.625} $\pm$ 0.125 & 0.940 $\pm$ 0.210 \\

        \midrule
        \multirow{5}{*}{\shortstack[l]{Vegetation\\+ Dirt}}
        & DBM                 & 0.061 $\pm$ 0.011 & 0.041 $\pm$ 0.009
                              & 0.276 $\pm$ 0.054 & 0.312 $\pm$ 0.070
                              & 0.824 $\pm$ 0.134 & 1.118 $\pm$ 0.255 \\

        & IKD FT-R            & 0.055 $\pm$ 0.010 & 0.033 $\pm$ 0.007
                              & 0.232 $\pm$ 0.045 & 0.452 $\pm$ 0.100
                              & 0.669 $\pm$ 0.127 & 1.452 $\pm$ 0.336 \\

        & AnyCar FT-R         & \textbf{0.054} $\pm$ 0.010 & \textbf{0.019} $\pm$ 0.004
                              & 0.207 $\pm$ 0.039 & 0.199 $\pm$ 0.045
                              & 0.776 $\pm$ 0.144 & 0.991 $\pm$ 0.221 \\

        \arrayrulecolor{black!85}\cdashline{2-8}[0.6pt/1.4pt]\arrayrulecolor{black}
        & \ourmethod{} FT-R   & 0.055 $\pm$ 0.011 & 0.020 $\pm$ 0.004
                              & 0.171 $\pm$ 0.033 & 0.172 $\pm$ 0.040
                              & 0.519 $\pm$ 0.108 & 0.981 $\pm$ 0.220 \\

        & \ourmethod{} FT-RS  & 0.059 $\pm$ 0.012 & 0.023 $\pm$ 0.005
                              & \textbf{0.157} $\pm$ 0.030 & \textbf{0.072} $\pm$ 0.016
                              & \textbf{0.348} $\pm$ 0.072 & \textbf{0.937} $\pm$ 0.224 \\

        \midrule
        \multirow{5}{*}{\shortstack[l]{Cart\\Pulling}}
        & DBM                 & 0.129 $\pm$ 0.025 & 0.014 $\pm$ 0.003
                              & 0.212 $\pm$ 0.041 & 0.197 $\pm$ 0.045
                              & 1.012 $\pm$ 0.163 & 1.881 $\pm$ 0.427 \\

        & IKD FT-R            & 0.067 $\pm$ 0.013 & 0.086 $\pm$ 0.019
                              & 0.244 $\pm$ 0.047 & 0.102 $\pm$ 0.023
                              & 0.711 $\pm$ 0.133 & 1.782 $\pm$ 0.403 \\

        & AnyCar FT-R         & 0.042 $\pm$ 0.008 & 0.031 $\pm$ 0.007
                              & 0.182 $\pm$ 0.035 & 0.049 $\pm$ 0.011
                              & 0.489 $\pm$ 0.095 & 1.338 $\pm$ 0.299 \\

        \arrayrulecolor{black!85}\cdashline{2-8}[0.6pt/1.4pt]\arrayrulecolor{black}
        & \ourmethod{} FT-R   & 0.028 $\pm$ 0.006 & 0.028 $\pm$ 0.006
                              & 0.174 $\pm$ 0.034 & 0.062 $\pm$ 0.014
                              & 0.319 $\pm$ 0.062 & 1.290 $\pm$ 0.294 \\

        & \ourmethod{} FT-RS  & \textbf{0.011} $\pm$ 0.002 & \textbf{0.014} $\pm$ 0.003
                              & \textbf{0.085} $\pm$ 0.017 & \textbf{0.044} $\pm$ 0.010
                              & \textbf{0.229} $\pm$ 0.054 & \textbf{1.023} $\pm$ 0.232 \\

        \bottomrule
    \end{tabular}
    \end{adjustbox}

    \caption{\textbf{Closed-loop figure-8 tracking with standard deviations.} Expanded version of Table~\ref{tab:fig8}, reporting TTE (m~$\downarrow$) and VTE (m/s~$\downarrow$) across terrains, speeds, and the unseen cart-pulling payload. Values report mean $\pm$ standard deviation of per-timestep tracking errors pooled across three runs.}
    \label{tab:fig8_std}
\end{table}

\end{document}